\documentclass[twoside,leqno,twocolumn]{article}

\usepackage[letterpaper]{geometry}

\usepackage{ltexpprt}
\usepackage{hyperref}

\usepackage{algorithm}
\usepackage{algorithmic}
\usepackage{amsmath}
\usepackage{amsfonts}
\usepackage{xspace}
\usepackage{graphicx}
\newcommand{\ours}{\texttt{MedDiffusion}\xspace}

\usepackage{multirow}
\usepackage{makecell}
\usepackage{colortbl}
\usepackage{color}
\usepackage{xcolor}
\usepackage{makecell}
\usepackage{subcaption}

\begin{document}

\newcommand\relatedversion{}
\renewcommand\relatedversion{\thanks{The full version of the paper can be accessed at \protect\url{https://arxiv.org/abs/1902.09310}}} % Replace URL with link to full paper or comment out this line

\title{\ours: Boosting Health Risk Prediction via Diffusion-based Data Augmentation}
\author{
    Yuan Zhong\textsuperscript{1},
    Suhan Cui\textsuperscript{1},
    Jiaqi Wang\textsuperscript{1},
    Xiaochen Wang\textsuperscript{1},
    Ziyi Yin\textsuperscript{1},\\
    Yaqing Wang\textsuperscript{2},
    Houping Xiao\textsuperscript{3},
    Mengdi Huai\textsuperscript{4},
    Ting Wang\textsuperscript{5},
    Fenglong Ma\textsuperscript{1} \\
    % \hfill\\
    \small \textsuperscript{1}The Pennsylvania State University,
    \textsuperscript{2}Purdue University,\\
    \small \textsuperscript{3}Georgia State University,
    \textsuperscript{4}Iowa State University,
    \textsuperscript{5}Stony Brook University \\
    \small \textsuperscript{1}{\{yfz5556, sxc6192, jqwang, zmy5171, xcwang, fenglong\}@psu.edu,
    \textsuperscript{2}wang5075@purdue.edu}\\
    \small \textsuperscript{3}hxiao@gsu.edu,
    \textsuperscript{4}mdhuai@iastate.edu,
    \textsuperscript{5}twang@cs.stonybrook.edu
}

\date{}

\maketitle

% Copyright Statement
% When submitting your final paper to a SIAM proceedings, it is requested that you include
% the appropriate copyright in the footer of the paper.  The copyright added should be
% consistent with the copyright selected on the copyright form submitted with the paper.
% Please note that "20XX" should be changed to the year of the meeting.

% Default Copyright Statement
% \fancyfoot[R]{\scriptsize{Copyright \textcopyright\ 2024 by SIAM\\
% Unauthorized reproduction of this article is prohibited}}

% Depending on which copyright you agree to when you sign the copyright form, the copyright
% can be changed to one of the following after commenting out the default copyright statement
% above.

%\fancyfoot[R]{\scriptsize{Copyright \textcopyright\ 20XX\\
%Copyright for this paper is retained by authors}}

%\fancyfoot[R]{\scriptsize{Copyright \textcopyright\ 20XX\\
%Copyright retained by principal author's organization}}

%\pagenumbering{arabic}
%\setcounter{page}{1}%Leave this line commented out.

\begin{abstract} \small\baselineskip=9pt Health risk prediction is one of the fundamental tasks under predictive modeling in the medical domain, which aims to forecast the potential health risks that patients may face in the future using their historical Electronic Health Records (EHR). Researchers have developed several risk prediction models to handle the unique challenges of EHR data, such as its sequential nature, high dimensionality, and inherent noise. These models have yielded impressive results. Nonetheless, a key issue undermining their effectiveness is data insufficiency. A variety of data generation and augmentation methods have been introduced to mitigate this issue by expanding the size of the training data set through the learning of underlying data distributions. However, the performance of these methods is often limited due to their task-unrelated design. To address these shortcomings, this paper introduces a novel, end-to-end diffusion-based risk prediction model, named \ours. It enhances risk prediction performance by creating synthetic patient data during training to enlarge sample space. Furthermore, \ours discerns hidden relationships between patient visits using a step-wise attention mechanism, enabling the model to automatically retain the most vital information for generating high-quality data. Experimental evaluation on four real-world medical datasets demonstrates that \ours outperforms 14 cutting-edge baselines in terms of PR-AUC, F1, and Cohen’s Kappa. We also conduct ablation studies and benchmark our model against GAN-based alternatives to further validate the rationality and adaptability of our model design. Additionally, we analyze generated data to offer fresh insights into the model's interpretability.\end{abstract}

% \footnote{The source code is available via: \url{https://shorturl.at/aerT0}.}

% \providecommand{\keywords}[1]{\textbf{\textit{KeyWords:}} #1}

\section{Introduction}

Predictive modeling in the healthcare domain aims to model patients' longitudinal electronic health records (EHR) with statistical and machine learning methods to identify disease-related patterns and predict task-related outcome probability. Among those predictive modeling tasks, the \textbf{health risk prediction task} is to forecast whether patients will develop or suffer from a disease-specific medical condition in the near future by modeling their EHRs. EHR data typically comprise patients’ time-ordered sequences of clinical visits, and each visit contains a collection of high-dimensional yet discrete medical codes such as the International Classification of Disease (ICD) codes. To model such unique data, researchers primarily adopt recurrent neural networks (RNN)~\cite{lstm} or Transformer~\cite{transformer} as backbones with advanced feature learning techniques, such as designing attention mechanisms~\cite{adacare,dipole,sand,retain} and modeling disease progression~\cite{retainex,tlstm} to further enhance the prediction performance. 

While existing models have demonstrated outstanding performance in health risk prediction, they face critical limitations of {data insufficiency}. These constraints arise from the relatively fixed population size and the low prevalence of certain diseases. In addition, privacy concerns further hinder access to comprehensive patient data globally, nationally, or even at a state level in the USA, while discrepancies between datasets from different healthcare organizations stifle the development of large-scale datasets and models. Furthermore, the under-representation of rare conditions within EHR data impedes the model's predictive capabilities. These limitations lead to a risk of overfitting when complex models and advanced techniques are applied to small EHR datasets.

% To address the aforementioned challenge of data insufficiency, \textbf{data augmentation} with Generative Adversarial Networks (GAN) become one of the most researched solutions to generate synthetic EHR data and effectively enlarge the medical data set. For instance, ~\cite{medGAN} generates synthetic data by learning patients' aggregated ICD code distribution, and ~\cite{ehrGAN} divides patients' visits by fixed 90-day windows and uses encoder-decoder Convolutional Neural Networks (CNN) structure to learn the latent distribution of EHR data. Another family of latent variable generation models, the Diffusion-based models, not only does excel in generation tasks~\cite{dalle2,nichol2021glide,pan2022synthesizing} but also shows prominent results as a data augmentation module in various tasks~\cite{aug_sound, aug_skeleton, aug_imagecaption,aug_image}. In the medical domain, this technique has been used for image augmentation on skin disease classification~\cite{aug_skin} and 3D brain magnetic resonance imaging (MRI) generation~\cite{aug_brain}. Among the emerging solutions, ~\cite{kotelnikov2023tabddpm} represents one of the first successful approaches, and is capable of generating tabular EHR data in both numerical and categorical forms, opening up new possibilities for medical data synthesis and providing a promising direction for overcoming the limitations observed in current techniques. However, state-of-the-art generation techniques primarily face the following drawbacks and cannot be directly applied to the health risk prediction task:

To address the aforementioned challenge of data insufficiency, \textbf{data augmentation} becomes one of the most researched solutions to generate synthetic EHR data and effectively enlarge the medical dataset. Generative Adversarial Networks (GAN)~\cite{goodfellow2020generative} and Diffusion-based models~\cite{ddpm} are two commonly-used approaches. 
For instance, MedGAN~\cite{medGAN} generates synthetic data by learning patients' aggregated ICD code distribution, and ehrGAN~\cite{ehrGAN} divides patients' visits by fixed 90-day windows and uses encoder-decoder Convolutional Neural Networks (CNN) structure to learn the latent distribution of EHR data. ~\cite{kotelnikov2023tabddpm} represents one of the first successful diffusion-based approaches and is capable of generating tabular EHR data in both numerical and categorical forms, opening up new possibilities for medical data synthesis and providing a promising direction for overcoming the limitations observed in current techniques. However, state-of-the-art generation techniques primarily face the following drawbacks and cannot be directly applied to the health risk prediction task:
\begin{itemize}
    % \item \textbf{Task-wise -- Target Irrelevance:} Existing EHR generation methods primarily focus on learning the distribution of existing EHR data; others train downstream prediction tasks as separate steps. \xiaochen{Not clear enough even though makes sense to me. } However, such approaches may not be optimal for generating task-specific EHR data, since during the data generation process, the interaction between the generation and the prediction module is ignored. An ideal solution would involve training an end-to-end prediction and generation model. In such a way, the downstream prediction module can act as guidance for generating task-augmented EHR data, while the generation module provides extra data diversity to boost the performance of the prediction module.\xiaochen{Better to put the final sentence in the beginning of this section, such as: Ideally, we should use prediction module for guidance such that ... however, previous works only focus on ... or ..., resulting in the ignorance of ...}
    \item \textbf{Task-wise -- Target Irrelevance:} Ideally, an \emph{end-to-end} prediction and generation model has the advantage that the risk prediction module can act as guidance in generating task-augmented EHR data. In turn, the generation module provides additional data diversity to boost the performance of the prediction module. However, existing work primarily concentrates on learning the distribution of existing EHR data and training the risk prediction task as separate steps. These strategies may not be the most effective for generating task-specific EHR data because the interaction between the generation and the prediction module is overlooked during the data generation process. Thus, their generated data may not be able to preserve and highlight the target-related information.
    % \item \textbf{Data-wise -- Data Manipulation:} Existing EHR generation methods also apply data manipulation techniques to aggregate patient data into one or several fix-windowed summarization vectors as the modeling input. However, existing work~\cite{hitanet,lsan,timeline} for health risk prediction proves that the visit dimension contains crucial information, such as disease progression, toward the risk prediction goal. Thus, direct modification or aggregation on the visit level would limit the usefulness of generated data. \xiaochen{Does not make sense to me... If visit dimension has been proven to be useful, why the summarization of multiple visits limit the usefulness?} 
    \item \textbf{Data-wise -- Data Manipulation:} By modeling the ordering and temporal relationship between visits, existing work~\cite{hitanet,lsan,timeline,tlstm} for the health risk prediction task has highlighted the significance of the visit dimension as a critical factor towards prediction. However, existing EHR generation methods often use data manipulation techniques that aggregate patient data into one or several fixed-window summarization vectors as the modeling input. Such approaches ignore the unique characteristics of EHR data, which inadvertently obscures valuable details such as disease progression from each patient's unique visit. Besides, aggregating by arbitrarily fixed windows has reduced the generalizability to other tasks and datasets.
    % \item \textbf{Model-wise -- Generation Architecture:} Although existing generation methods have achieved notable progress in patient-level data generation, it is hard to directly apply to visit-level data generation since the generated data might occasionally suffer from intra-visit misalignment, which could potentially impact the downstream predictive performance. Therefore, incorporating mechanisms that effectively facilitate the learning process and preserve visit-level information in the data augmentation module of the new end-to-end health risk prediction model becomes crucial.
\end{itemize}

\begin{figure*}[]
    \centering
    \includegraphics[width=0.90\textwidth]{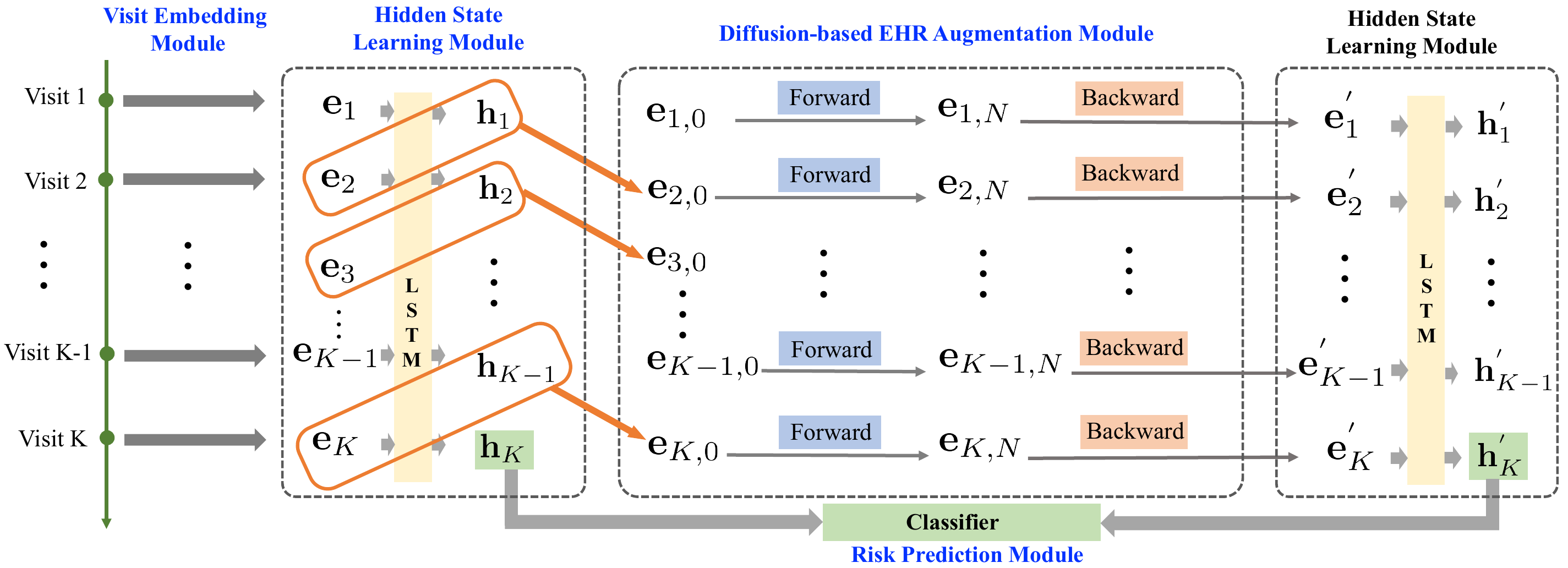}
    \vspace{-2mm}
    \caption{Overview of the proposed {\ours} model.}
    \label{mainModel}
\end{figure*}

To address these drawbacks simultaneously, in this paper, we propose a novel \textbf{end-to-end} health risk prediction model named {\ours} with a special diffusion-based data augmentation module, as shown in Figure~\ref{mainModel}. 
It consists of four components: the visit embedding module, the hidden state learning module, the diffusion-based EHR augmentation module, and the risk prediction module. The \emph{visit embedding module} aims to map each visit into a vector representation $\mathbf{e}_k$, where each visit is associated with a set of diagnosis codes (i.e., $v_k$) and time information $t_k$. In the \emph{hidden state learning module}, a long short-term memory (LSTM) network takes the time-ordered visit embeddings $[\mathbf{e}_1, \cdots, \mathbf{e}_K]$ as the input to generate the corresponding hidden states $[\mathbf{h}_1, \cdots, \mathbf{h}_K]$.
Both $[\mathbf{e}_1, \cdots, \mathbf{e}_K]$ and $[\mathbf{h}_1, \cdots, \mathbf{h}_K]$ are the inputs of the \emph{diffusion-based EHR augmentation module}. \ours take both the current visit embedding $\mathbf{e}_k$ and the previous visit information represented by the hidden state $\mathbf{h}_{k-1}$ into consideration when generating the synthetic visit $\mathbf{e}_k^\prime$. In particular, we propose a new step-wise attention mechanism to aggregate $\mathbf{e}_k$ and $\mathbf{h}_{k-1}$ to make the diffusion model computable.
Finally, in the \emph{risk prediction module}, we consider two predictions from the original EHR data and the generated EHR data when optimizing \ours. 

To sum up, our \textbf{contributions} are listed as follows: (1) To the best of our knowledge, this is the first work to augment time-ordered yet discrete EHR data through diffusion-based methods for the healthcare risk prediction task in the medical domain. (2) We propose a novel data augmentation module {\ours} that generates synthetic EHR data in the continuous space and takes the inner relationships among visits into account during the generation. (3) We conduct experiments on both private and public real-world medical datasets to demonstrate the effectiveness of the proposed {\ours} model compared with state-of-the-art baselines, and model insight analysis shows the reasonableness and generalizability of our model design.
\section{Related Work}\label{sec:related}
\subsection{Health Risk Prediction and EHR Generation}
Since the EHR data is ordered sequences, many of the existing studies are built on sequential models such as RNN and Transformer~\cite{transformer}, and utilize various attention mechanisms on either local or global levels to highlight important information~\cite{dipole, retain, retainex, tlstm, timeline,sand, hitanet, medskim, lsan}. Along with visits and codes weighting methods, augmenting medical data through extra knowledge~\cite{ma2018kame,ye2021medpath,chen2021unite} is also a popular strategy to mitigate the effect of sample selection bias that training data does not sufficiently represent the real-world scenario or the data itself does not include sufficient information towards the prediction goal. On the other hand, medical data generation aims to generate synthetic medical data in either numerical or discrete forms~\cite{actGAN,GcGAN,ehrGAN,medGAN,kotelnikov2023tabddpm} to alleviate data scarcity or privacy concerns in the health domain. However, existing generation methods are not task-related and fail to address the unique characteristics of EHR data and thus are suboptimal in their performance.

\subsection{Diffusion-based Generation and Augmentation}
The denoising diffusion probabilistic model (DDPM) is a representative diffusion-based model and has shown great success in many tasks. It has been used for continuous data generation such as the image generation tasks~\cite{li2022srdiff,rombach2022high,saharia2022image,dalle2,nichol2021glide,popov2021grad} and time series forecasting or imputation~\cite{rasul2021autoregressive,tashiro2021csdi}. The discrete data generation tasks also employ DDPM, by transforming discrete tokens into continuous embeddings and mapping back after generation~\cite{austin2021structured,li2022diffusionlm,gong2022diffuseq}. In the meantime, many studies have proposed integrating DDPM into different domain-specific models as a data augmentation tool~\cite{aug_image,aug_camdiff,aug_imagecaption,aug_skeleton,aug_sound}. Specifically in the medical domain, researchers have started their first attempt to utilize DDPM to generate EHR data~\cite{kotelnikov2023tabddpm}. However, existing methods of health data generation do not consider the temporal relationship within EHR data and thus cannot be directly applied to the health risk prediction task. 
\section{Methodology}\label{sec:model}
The EHR data consists of a list of patients' information collected by healthcare providers, including the date and time of each visit and medication codes for any diseases or conditions of patients. 
Let $V = [(v_1,t_1),(v_2,t_2),\dots,(v_K,t_K)]$ denote a patient's visit data, where $K$ is the total number of visits, and $t_k$ is the timestamp of the $k$-th visit. 
Each visit $v_k$ contains a set of unordered ICD codes.  Let $v_k = [c_1^k,c_2^k,\dots,c_M^k]$ denote the binary vector representation of the codes appearing in visit $v_k$, where $M$ represents the total number of unique ICD codes in the dataset. $c_i^k = 1$ means the $i$-th code that appears in $v_k$; otherwise $c_i^k = 0$.
Each patient is also associated with a label $y\in \{1,0\}$, representing whether the patient is a positive or negative case for the target disease. The \textbf{task} of health risk prediction is to predict whether a particular patient will develop a specific disease or condition in the future by analyzing the historical EHR data $V$. 
The proposed {\ours} model consists of four modules, as shown in Figure~\ref{mainModel}. Next, we will introduce the detailed design of each module one by one.

\subsection{Visit Embedding}
Following existing work~\cite{hitanet,medskim}, we first map each visit to a vector representation consisting of two embeddings. One is the diagnosis code embedding $\mathbf{v}_k$, and the other is the time embedding $\mathbf{t}_k$. Let $\hat{\mathbf{e}}_k$ be the visit embedding, and we have
\begin{equation}\label{eq:visit_embedding}
    \mathbf{e}_k = \mathbf{v}_k + \mathbf{t}_k.
\end{equation}
The visit embedding $\mathbf{v}_t$ is calculated as follows:
\begin{equation}
    \mathbf{v}_k = \text{ReLU}(\mathbf{W}_v v_k + \mathbf{b}_v),
    \label{codeEmbedding}
\end{equation}
where $\mathbf{W}_v\in \mathbb{R}^{d_e \times M}$ and \(\mathbf{b}_v\in \mathbb{R}^{d_e}\) are learnable parameters.
Following~\cite{hitanet}, we use the time gap $\Delta t_k$ between the last time $t_K$ and the current visit time $t_k$ (i.e., $\Delta t_k = t_K - t_k$) to model the time embedding as follows:
\begin{equation}
    \mathbf{t}_k = \mathbf{W}_t \left(1-\text{Tanh}\left((\mathbf{W}_{f}\frac{\Delta t_k}{180}+\mathbf{b}_f)^2\right)\right) + \mathbf{b}_t,
    \label{timeEmbedding}
\end{equation}
where \(\mathbf{W}_f\in \mathbb{R}^{d_f}\), \(\mathbf{b}_f\in \mathbb{R}^{d_f}\), \(\mathbf{W}_t\in \mathbb{R}^{d_e \times d_f}\), and \(\mathbf{b}_t\in \mathbb{R}^{d_e}\) are learnable parameters.

\subsection{Hidden State Learning}

Using the obtained visit embeddings $[\mathbf{e}_1, \cdots, \mathbf{e}_K]$ via Eq.~\eqref{eq:visit_embedding}, we can apply an RNN model such as LSTM to generate hidden states as follows: 
\begin{equation}\label{eq:hidden_states}
    [\mathbf{h}_1, \cdots, \mathbf{h}_K] = \text{LSTM}([\mathbf{e}_1, \cdots, \mathbf{e}_K]),
\end{equation}
where $\mathbf{h}_k \in \mathbb{R}^{d_h}$ is the $k$-th hidden state. 

\subsection{Diffusion-based EHR Data Augmentation}\label{sec:diffusion_model}

To further enhance the learning capacity of LSTM, we propose to augment EHR data based on the Denoising Diffusion Probabilistic Model (DDPM)~\cite{ddpm}. DDPM mainly contains two components, i.e., the forward diffusion process and the backward inference process. 
Most of the existing DDPM techniques are mainly used to generate images in continuous space~\cite{saharia2022image,rombach2022high,li2022srdiff}. However, medical data are significantly different from image data. Medical data can be considered as time-ordered sequences. The information on the current visit may be highly related to that of the previous ones, which requires the augmentation model to have the ability to take previous information as input. Toward this end, we propose a new diffusion-based model for augmenting time-ordered EHR data. In particular, to avoid introducing mapping errors, we propose to augment latent representations of visits in continuous space instead of discrete medical codes.

\subsubsection{Forward Diffusion Process}
The forward diffusion process aims to add noise to the input data gradually. Let $N$ denote the number of steps, and the noise of each step is drawn for a Gaussian distribution. Our goal is to generate a sequence of visits based on the input visit embeddings $[\mathbf{e}_1, \cdots, \mathbf{e}_K]$ learned by Eq.~\eqref{eq:visit_embedding}. As we discussed before, the generation of each visit $\mathbf{e}_k$ should take the previous information $[\mathbf{e}_1, \cdots, \mathbf{e}_{k-1}]$ into consideration. Mathematically, the forward diffusion process is defined as follows:
\begin{equation}
\begin{aligned}
      q(\mathbf{e}_{k, 1:N}| \mathbf{e}_{k,0},& \mathbf{e}_{k-1}, \cdots,\mathbf{e}_{1}) = \\ &\prod_{n=1}^N q(\mathbf{e}_{k,n}|\mathbf{e}_{k,n-1}, \mathbf{e}_{k-1},\cdots,\mathbf{e}_{1}),  
\end{aligned}
\end{equation}
where $\mathbf{e}_{k,0}$ is the $k$-th visit embedding learned by Eq.~\eqref{eq:visit_embedding}. As a Markov Chain, every step of the forward diffusion process is a Gaussian distribution that only depends on its previous step. Thus, we can further rewrite the forward process into discrete steps by gradually adding noise to the intermediate noised data $\mathbf{\mu}$ in $N$ steps according to the noise schedule $\beta_n\in (0,1)_{n=1}^N$ as follows: 
\begin{equation}
\begin{aligned}
q(&\mathbf{e}_{k,n}|\mathbf{e}_{k,n-1},\mathbf{e}_{k-1},\cdots,\mathbf{e}_{1})= \\
 &\mathcal{N}(\mathbf{e}_{k,n};\sqrt{1-\beta_n}
\mathbf{\mu}(\mathbf{e}_{k,n-1},\mathbf{e}_{k-1},\cdots,\mathbf{e}_1),\beta_n\mathbf{I}). 
\end{aligned}
\end{equation}

\subsubsection{Backward Diffusion Process}
The backward process, on the other hand, aims to obtain the original input data from pure Gaussian samples, i.e., reconstructing $\mathbf{e}_{k,0}$ using $\mathbf{e}_{k,N}$ and previous information $[\mathbf{e}_1,\cdots,\mathbf{e}_{k-1}]$. Thus, our backward process can be written as follows:
\begin{equation}
\begin{aligned}
p_\theta(\mathbf{e}_{k,0:N})= 
p&(\mathbf{e}_{k,N},\mathbf{e}_{k-1},\cdots,\mathbf{e}_{1})\\&\prod_{n=1}^{N}p_\theta(\mathbf{e}_{k,n-1}|\mathbf{e}_{k,n},\mathbf{e}_{k-1},\cdots,\mathbf{e}_{1}),
\end{aligned}
\label{backward1}
\end{equation}
where $\theta$ is the parameter set in the diffusion model.
Similar to the forward process, each step of the backward process can be represented as a Gaussian distribution with the approximated mean and fixed variance as follows:
\begin{equation}
\begin{aligned}
p_\theta(&\mathbf{e}_{k,n-1}|\mathbf{e}_{k,n},\mathbf{e}_{k-1},\cdots,\mathbf{e}_{1}) = \\
\mathcal{N} (&\mathbf{e}_{k,n-1};\sqrt{1-\Tilde{\beta}_n}\mathbf{\Tilde{\mu}}(\mathbf{e}_{k,n},\mathbf{e}_{k-1},\cdots,\mathbf{e}_{1}),\Tilde{\beta}_n\mathbf{I}),  
\end{aligned}
\label{backward2}
\end{equation}
where $\mathbf{\Tilde{\mu}}$ and $\Tilde{\beta}_n$ are approximated mean and predefined noise schedule, respectively.

As shown in Eq.~\eqref{backward1}, we not only need to calculate the conditional probability of the diffusion step itself but also include the conditional probabilities between all previous visits and the current visit. It makes the current backward diffusion process  impractical due to the complex computation. To solve this problem, we need to take one step back and reformulate Eq.~\eqref{backward1}.

To be specific, we relax the condition constraints on all previous visits $[\mathbf{e}_{k-1},\cdots,\mathbf{e}_1]$ in Eq.~\eqref{backward1} and make an assumption that the learned hidden state of previous visits $\mathbf{h}_{k-1}$ from LSTM is sufficient of accumulating information from all previous visits. Thus, we can replace $[\mathbf{e}_{k-1},\cdots,\mathbf{e}_1]$ with $\mathbf{h}_{k-1}$ in the backward process and simplify Eq.~(\ref{backward1}) as follows:
\begin{equation}
\begin{aligned}
p_\theta(\mathbf{e}_{k,0:N})
= 
p(\mathbf{e}_{k,N}&,\mathbf{h}_{k-1})\\ &\prod_{n=1}^{N}p_\theta(\mathbf{e}_{k,n-1}|\mathbf{e}_{k,n},\mathbf{h}_{k-1}). 
\end{aligned}
\label{backward3}
\end{equation}

Consequently, the Gaussian steps of the backward process can also be written with $\mathbf{h}_{k-1}$ as follows:
\begin{equation}
\begin{aligned}
p_\theta(\mathbf{e}_{k,n-1}&|\mathbf{e}_{k,n},\mathbf{h}_{k-1}) = \\
\mathcal{N} &(\mathbf{e}_{k,n-1};\sqrt{1-\Tilde{\beta}_n}\Tilde{\mu}(\mathbf{e}_{k,n},\mathbf{h}_{k-1}),\Tilde{\beta}_n\mathbf{I}).
\end{aligned}
\label{backward4}
\end{equation}

\subsubsection{Step-wise Information Aggregation via Attention}\label{sec:step-wise-attention}
Another issue that we are facing in solving Eq.~\eqref{backward4} is that the generated $\mathbf{e}_{k,n}$ and the hidden state $\mathbf{h}_{k-1}$ are not in the same latent space. Thus, we cannot mandatorily force them together. To make Eq.~\eqref{backward4} computable, we need first to map them to the same space. Toward this end, we propose to use a step-wise attention mechanism to automatically distinguish the influence of the visit and the hidden state for the generation as follows:
\begin{equation}
\begin{aligned}
      [\gamma^{e}_n, \gamma^{h}_n] &= \\
      \text{Softmax} & \left( \mathbf{W}_a(\text{Tanh}(\mathbf{W}_b 
    \begin{bmatrix}
\mathbf{e}_{k,n}\\
\mathbf{W}_h\mathbf{h}_{k-1}
\end{bmatrix}
    +\mathbf{b}_b))\right)  
\end{aligned}
\label{fuse}
\end{equation}
where $\mathbf{W}_a \in \mathbb{R}^{2 \times d_b}$, $\mathbf{W}_b \in \mathbb{R}^{d_b \times 2d_e}$, $\mathbf{W}_h\in \mathbb{R}^{d_e \times d_h}$ and $b_b\in \mathbb{R}^{d_b}$ are learnable parameters. $\begin{bmatrix}
\cdot\\
\cdot
\end{bmatrix}$ is the concatenation operation. $\gamma^e_n$ is the attention weight for the generated noise $\mathbf{e}_{k,n}$, and $\gamma^h_n$ is the weight for the hidden state $\mathbf{h}_{k-1}$. Let $\hat{\mathbf{e}}_{k,n} = \gamma^e_n\mathbf{e}_{k,n}+\gamma^h_n \mathbf{W}_h\mathbf{h}_{k-1}$, and we can rewrite Eq.~\eqref{backward4} as follows:
\begin{equation}
\begin{aligned}
p_\theta(\mathbf{e}_{k,n-1}&|\mathbf{e}_{k,n},\mathbf{h}_{k-1}) = \\
&\mathcal{N} \left(\mathbf{e}_{k,n-1};\sqrt{1-\Tilde{\beta}_n}\Tilde{\mu}(\hat{\mathbf{e}}_{k,n}),\Tilde{\beta}_n\mathbf{I}\right).
\end{aligned}
\label{backwardgamma}
\end{equation}

With the above calculation, we produce synthetic data sequence $[\mathbf{e}_1^{\prime},\mathbf{e}_2^{\prime},\cdots,\mathbf{e}_K^{\prime}]$ for each visit per patient. We then use the same LSTM to generate the hidden states $[\mathbf{h}_1^{\prime},\mathbf{h}_2^{\prime},\cdots,\mathbf{h}_K^{\prime}]$ via Eq.~\eqref{eq:hidden_states} for the generated patient's data. 

\subsection{Risk Prediction}
Since there are two sets of hidden states, one is calculated from the original EHR data, and the other is from the generated data, we can make predictions using both of them.

The last hidden state $\mathbf{h}_K$ learned by Eq.~\eqref{eq:hidden_states} can be used to predict the risk as follows:
\begin{equation}\label{eq:prediction}
    \hat{\mathbf{y}} = \text{Softmax}(\mathbf{W}_y \mathbf{h}_K + \mathbf{b}_y),
\end{equation}
where $\hat{\mathbf{y}} \in \mathbb{R}^2$ is the prediction probability vector, $\mathbf{W}_y \in \mathbb{R}^{2 \times d_h}$, and $\mathbf{b}_y \in \mathbb{R}^2$ are parameters. 
Similarly, we can use the last hidden state $\mathbf{h}^\prime_K$ generated by the synthetic data to make a prediction as Eq.~\eqref{eq:prediction}, i.e., 
\begin{equation}\label{eq:prediction_fake}
    \hat{\mathbf{y}}^\prime = \text{Softmax}(\mathbf{W}_y \mathbf{h}_K^\prime + \mathbf{b}_y).
\end{equation}

\subsection{Loss Function}
The final loss of the proposed {\ours} model consists of three parts as follows:
\begin{equation}
    \mathcal{L} = \mathcal{L}_{\text{LSTM}} + \lambda_S \mathcal{L}_{\text{LSTM}}^\prime + \lambda_D \mathcal{L}_{\text{Diffusion}},
    \label{totalloss}
\end{equation}
where $\mathcal{L}_{\text{LSTM}}$ is the loss from the original data, $\mathcal{L}_{\text{LSTM}}^\prime$ denotes the loss from the generated data, and $\mathcal{L}_{\text{Diffusion}}$ is the loss from the diffusion model. $\lambda_S$ and $\lambda_D$ are the hyperparameters to balance these losses.

The cross-entropy (CE) loss can be used to optimize the risk prediction model as follows:
\begin{equation}\label{eq:lstm_real}
    \mathcal{L}_{\text{LSTM}} = \sum_{j = 1}^J\text{CE}({\mathbf{y}_j}, \hat{\mathbf{y}}_j),
\end{equation}
where $J$ is the number of training data, $\mathbf{y}_j$ is the ground truth one-hot vector for the $j$-th data, and $\hat{\mathbf{y}}_j$ is the $j$-th data's prediction vector learned by Eq.~\eqref{eq:prediction}.
Similar to Eq.~\eqref{eq:lstm_real}, we can calculate the loss from the generated data using
$\mathcal{L}_{\text{LSTM}}^\prime = \sum_{j = 1}^J\text{CE}({\mathbf{y}_j}, \hat{\mathbf{y}}_j^\prime)$, where $\hat{\mathbf{y}}_j^\prime$ is the prediction using Eq.~\eqref{eq:prediction_fake}.

The diffusion model is trained to minimize the negative log-likelihood $\mathbb{E}[-\log p_\theta(\hat{\mathbf{e}}_{k,0})]$, which can be obtained with the variational lower bound $\mathbb{E}_q[-\log\frac{p_\theta(\hat{\mathbf{e}}_{k,0:N})}{q(\hat{\mathbf{e}}_{k,1:N}|\hat{\mathbf{e}}_{k,0})}]$ and written in terms of the sum of Kullback–Leibler divergence as follows:
\begin{equation}
\begin{aligned}
\mathcal{L}(\hat{\mathbf{e}}_{k,0}) = \mathbb{E}_q[\log&\frac{q(\hat{\mathbf{e}}_{k,N}|\hat{\mathbf{e}}_{k,0})}{p_\theta(\hat{\mathbf{e}}_{k,N})} -\log p_\theta(\hat{\mathbf{e}}_{k,0}|\hat{\mathbf{e}}_{k,1})  \\ +
&\sum_{n=2}^{N}\log\frac{q(\hat{\mathbf{e}}_{k,n-1}|\hat{\mathbf{e}}_{k,n-1},\hat{\mathbf{e}}_{k,0})}{p_\theta(\hat{\mathbf{e}}_{k,n-1}|\hat{\mathbf{e}}_{k,n})}].   
\end{aligned}
\label{unstableloss}
\end{equation}

As stated in DDPM~\cite{ddpm}, the above optimization objective is unstable and hard to optimize. We then shift from reconstructing the input sample $\hat{\mathbf{e}}_{k,0}$ to learn the amount of noise that needs to be deleted from the Gaussian noise sample $\hat{\mathbf{e}}_{k,N}$. 
Thus, we follow the simplification procedure and derive the loss function $\mathcal{L}_{\text{Diffusion}}$ to learn the added noise of one visit $\hat{\mathbf{e}}_{k,0}$ and aggregate alone visit dimension $K$ as follows:
\begin{equation}
\mathcal{L}_{\text{Diffusion}}=\sum_{k=1}^{K}\sum_{n=1}^{N}\mathbb{E}_{q}(\epsilon-\epsilon_{\theta}(\hat{\mathbf{e}}_{k,0},n))^2,
    \label{diffusionLoss}
\end{equation}
where $\epsilon$ is the noise of closed form Gaussian posterior, and $\epsilon_\theta$ is the predicted noise of the neural network.
\begin{table}[t]
\caption{Statistics of datasets.}
\vspace{-0.1in}
\centering
\resizebox{0.95\columnwidth}{!}{
\begin{tabular}{l|ccc|c}
\hline
\textbf{Dataset}           & \textbf{Kidney}  & \textbf{COPD} & \textbf{Amnesia}     & \textbf{MIMIC} \\ \hline
Positive Cases     & 2,810    & 7,314   & 2,982          &  2,820\\
Negative Cases        & 8,430     & 21,942  & 8,946            &  4,702\\
Average Visits per Patient & 39.09  & 30.39   & 39.00         &  2.61\\
Average Code per Visit    & 4.70  & 3.50  & 2.53         &  13.06\\
Unique ICD-9 Codes        & 8,802   & 10,053   & 9,032      &  4,874  \\ \hline
\end{tabular}
}
\label{datasetTable}
\end{table}

\section{Experiments}\label{sec:exp}
In this section, we conduct experiments on four real-world datasets to validate the effectiveness of \ours compared with state-of-the-art baselines. \emph{Note that we put algorithm flow and extra experimental results in the appendix, including hyperparameter study, hidden state analysis, and module analysis.}

\begin{table*}[ht]
\caption{Performance comparison in terms of PR-AUC, F1, and Cohen's Kappa on the four datasets.}
\vspace{-0.1in}
\centering
\resizebox{0.98\textwidth}{!}{
\begin{tabular}{c|c|ccc|ccc|ccc|ccc}
\hline
\multirow{2}{*}{\textbf{Category}}& \textbf{Datasets}  & \multicolumn{3}{c|}{\textbf{Kidney}}  & \multicolumn{3}{c|}{\textbf{COPD}}  & \multicolumn{3}{c|}{\textbf{Amnesia}} & \multicolumn{3}{c}{\textbf{MIMIC}}  \\ \cline{2-14}
&Metrics    & PR-AUC   & F1      & Kappa    & PR-AUC   & F1      & Kappa   & PR-AUC   & F1      & Kappa & PR-AUC   & F1      & Kappa \\ \hline
\multirow{11}{*}{\thead{Health\\ Risk\\ Prediction}}
&LSTM    & 61.07   & 63.50  & 50.95  & 55.34  & 55.96 & 41.78 & 53.63 & 60.97 & 47.67   & 59.43 & 57.58 & 34.61 \\
&Dipole & 65.33   & 60.71  & 48.63  & 58.70  & 56.18 & 42.18 & 57.38 & 61.83 & 53.15 & 58.56 & 57.40 & 33.49\\
&Retain  & 57.81   & 57.25  & 44.14  & 53.56  & 50.96 & 37.46  & 59.51 & 54.98 & 43.95 & 59.89 & 59.13 & 37.20\\
&SAnD    & 54.65   & 60.23  & 43.14  & 51.70  & 52.12 & 37.66  & 53.30 & 58.42 & 44.19  & 54.70 & 54.51 & 30.87\\ 
&Adacare & 69.52   & 62.58  & 49.66   & 60.50  & 55.08 & 42.34 & 59.36 & 60.23 & 47.56 & 62.42 & 61.36 & 36.27\\
&LSAN     & 73.20   & 65.36  & 53.34  & 63.84  & 54.98 & 43.52 & 68.85 & 64.00 & 53.56 & 69.01 & 66.35 & 38.98\\
&RetainEx & 69.57   & 62.40  & 50.57  & 60.52  & 54.04 & 43.44 & 65.55 & 59.61 & 50.24 & 61.52 & 58.23 & 35.78\\
&Timeline & 68.07   & 60.45  & 48.48 & 54.86  & 49.02 & 36.40 & 57.14 & 57.71 & 45.00 & 65.45 & 60.49 & 39.53\\
&T-LSTM   & 69.40 & 67.39  & 55.87  & 68.62  & 62.92 & 51.55 & 60.42 & 61.64 & 49.34 & 61.93 & 61.16 & 38.86\\
&HiTANet  & 75.54  & 68.85  & 57.23 & 68.46  & 63.70 & 51.78 & 69.30 & 63.17 & 51.39 & 60.44 & 60.78 & 37.38\\
&MedSkim & 76.31   & 68.58  & 57.07  & 69.32  & 63.72 & 52.01 & 70.85 & 65.26 & 53.67 & 62.20 & 61.44 & 37.06\\ \hline
\multirow{3}{*}{\thead{EHR\\ Augmentation}}
&MaskEHR & 70.08 & 62.94 & 50.74 & 61.16 & 50.92 & 39.54 & 68.58 & 57.66& 47.50 & 59.42 & 58.90 & 36.72\\
&ehrGAN & 55.80 & 57.40 & 41.50 & 45.38 & 47.98 & 29.15 & 54.69 & 60.06 & 45.44 & 43.46 & 58.12 & 22.02 \\
&TabDDPM & 60.30 & 56.36 & 48.99 & 57.54 & 57.08 & 36.40 & 56.16 & 57.64 & 42.35 & 56.20 & 62.67 & 39.36\\\hline
Ours & \ours & \textbf{77.88}   & \textbf{70.36}  & \textbf{58.82} & \textbf{72.03}  & \textbf{65.26} & \textbf{54.21}  & \textbf{74.69}   & \textbf{68.43}  & \textbf{57.42}   & \textbf{70.64} & \textbf{66.79} & \textbf{45.26}\\ \hline

\end{tabular}
}
\label{resultTable1}
\end{table*}

\subsection{Implementation}
Our model is implemented in PyTorch and trained on an NVIDIA RTX A6000 GPU. We use the Adam optimizer with learning rate and weight decay both set to \(10^{-3}\), and employ a ReduceLROnPlateau scheduler with patience 5 and factor 0.2. The dimensions for the various modules are \(d_f = 64\), \(d_e = d_h = 256\), and \(d_b = 64\). Loss function hyperparameters are \(\lambda_D = 0.1\) and \(\lambda_S = 0.5\). Baselines are run under the same settings. The dataset is divided as 75\% for training, 10\% for validation, and 15\% for testing. Model selection is based on the F1 score on the validation set and averaged over 5 runs.

\subsection{Experimental Setup}\hfill\\
\textbf{Datasets}.
Four chronic and progressive health conditions are chosen to conduct a retrospective analysis. Three of them are from TriNetX (COPD, Amnesia, Kidney disease) and one is from MIMIC-III~\cite{johnson2016mimic} (Heart Failure). Following the data preprocessing procedure described in~\cite{retain} that is carried on to various state-of-the-art health risks prediction models~\cite{medskim,hitanet}, the data are extracted from under the guidance of clinicians and then re-formatted into a time series format starting from six months before the first diagnosis date for each patient. Three control cases are chosen for each positive case based on matching criteria such as gender, age, race, and underlying diseases. Table~\ref{datasetTable} presents statistics of the four datasets used in our experiments.

\noindent\textbf{Baselines}.
In this section, we present the following state-of-the-art \textbf{health risk prediction} models~\cite{lstm,dipole,retain,adacare,timeline,retainex,tlstm,medskim,sand,lsan,hitanet} as baselines, as well as three \textbf{EHR augmentation} models~\cite{ma2020rare,ehrGAN,kotelnikov2023tabddpm}. Detailed explanations of each baseline model are in the appendix.

\noindent\textbf{Evaluation Metrics}.
Since the datasets used in the experiments are imbalanced, as shown in Table~\ref{datasetTable}, we choose the following three metrics in percentage to evaluate our models' performance: (1) the Area Under the Precision-Recall Curve (\textbf{PR-AUC}), (2) \textbf{F1 score}, and (3) \textbf{Cohen's Kappa score}, which are widely-used to evaluate the imbalanced data.

% The proposed model is implemented in Pytorch on an NVIDIA RTX A6000 GPU. The parameters are trained with the Adam optimizer with the initial learning rate of 10e-3 and weight decay of 10e-3. The learning rate scheduler ReduceLROnPlateau is also used with parameters patient set to 5 and factor set to 0.2. In the visit embedding learning module, the intermediate layer dimension $d_f$ is set to 64. Both visit embedding dimension $d_e$ and hidden state dimension $d_h$ of LSTM are set to 256. As a result, the diffusion module's dimension is the same as $d_{e}$. In the information aggregation module, we set the intermediate layer dimension $d_b$ to 64. For the hyperparameters in the loss function, we set the regularization term $\lambda_D$ to 0.1, and $\lambda_S$ to 0.5. The baselines are implemented through the same platform and suggested parameters in their original settings. The data sets are randomly partitioned into training, validation, and test set, with a ratio of 0.75:0.10:0.15. The model is selected by the performance on the validation set by the F1 score, and for each model, we ran 5 times to get the mean of performance metrics as the experimental results.

\subsection{Performance Evaluation}
% Table \ref{resultTable1} shows the experimental results of all baselines and the proposed model \ours on Kidney, COPD, Amnesia, and MIMIC datasets. All models are run \textbf{five times}, and the average values of PR-AUC, F1, and Kappa are presented in the result table. We can observe that the proposed \ours achieves the overall best performance in all settings. 
Table \ref{resultTable1} presents averaged PR-AUC, F1, and Kappa values from \textbf{five runs} across four datasets. Our proposed model \ours consistently outperforms all baselines.

Comparing the proposed model \ours against the backbone LSTM, we can observe a significant performance increase, e.g., in terms of PR-AUC, a 27.5\% increase on the Kidney dataset and a 30.2\% increase on the COPD dataset. This is because the plain LSTM does not model time gaps between visits and cannot highlight important visits, which leads to less strength against information decay. Even on the MIMIC dataset with less average per visit, our model still achieves an 18.5\% increase. For models that incorporate basic attention mechanisms, Dipole and Retain do not consider time information and mainly weight visits or ICD codes by attention, resulting in a limited performance increase. SAnD has lower performance across all data sets against LSTM, possibly caused by the unstableness introduced by interpolation. Advanced models like AdaCare, RetainEx, and Timeline outperform simpler ones by incorporating time-aware attention mechanisms. LSAN and HiTANet incorporate transformer-based hierarchical attention, while Medskim focuses on selecting the most relevant visits and codes. However, these models are still limited by the quality and noise in the training data. MaskEHR's attempt to augment rare category EHR data suffers from mapping errors, affecting its performance. ehrGAN's arbitrary 90-day aggregation window causes information loss, and TabDDPM underperforms as it's tailored for simpler, tabular data.

Unlike all baselines, our proposed model takes the direction of data augmentation on the embedding space and achieves a consistent performance increase compared with the best-performing models. With step-wise information aggregation, we magnify the temporal and visit relationships between visits. With the diffusion module, we can reliably generate data on learning embedding space as augmentation. Furthermore, since the data augmentation happens in the embedding space, we do not need to face the extra noise caused by the rounding step from embedding to codes as in Diffusion-LM~\cite{li2022diffusionlm}. Thus, our model can perform better than baselines.

% \begin{table}[t]
% \caption{Performance comparison in terms of PR-AUC, F1, and Cohen's Kappa on MIMIC-III.}
% % \vspace{0.1in}
% \centering
% \resizebox{0.9\columnwidth}{!}{
% \begin{tabular}{c|ccc}
% \hline
% Datasets  & \multicolumn{3}{c}{MIMIC-III} \\ \hline
% Metrics    & PR-AUC   & F1      & Kappa  \\ \hline
% LSTM    & 59.43 & 57.58 & 34.61 \\
% Dipole  & 58.56 & 57.40 & 33.49\\
% Retain  & 59.89 & 59.13 & 37.20\\
% SAnD    & 54.70 & 54.51 & 30.87\\ 
% Adacare & 62.42 & 61.36 & 36.27\\
% LSAN     & 69.01 & 66.35 & 38.98\\
% RetainEx & 61.52 & 58.23 & 35.78\\
% Timeline & 65.45 & 60.49 & 39.53\\
% T-LSTM   & 61.93 & 61.16 & 38.86\\
% HiTANet  & 60.44 & 60.78 & 37.38\\
% MedSkim  & 62.20 & 61.44 & 37.06\\ 
% MaskEHR  & 59.42 & 58.90 & 36.72\\
% ehrGAN  & 43.46 & 58.12 & 22.02 \\
% TabDDPM  & 56.20 & 62.67 & 39.36\\
% \ours & \textbf{70.64} & \textbf{66.79} & \textbf{45.26}\\ \hline
% \end{tabular}
% }
% \label{resultTable2}
% \vspace{-0.2in}
% \end{table}

\subsection{Ablation Study}

% \begin{table}[h]
% \centering
% \caption{Ablation study results in terms of PR-AUC.}
% % \vspace{-0.1in}
% \resizebox{0.9\columnwidth}{!}{
% \begin{tabular}{l|cccc|c}
% \hline
% Dataset  & Kidney   & COPD  & Amnesia & Heart Failure & MIMIC \\ \hline
% AS-1 & 76.91 & 70.19 & 71.90  & 70.11 & 67.98\\
% AS-2 & 77.18 & 71.44 & 72.90 & 70.84 & 68.42\\
% AS-3 & 76.82 & 71.00 &  69.17  & 69.63 & 65.06\\ 
% \ours & \textbf{77.88} & \textbf{72.03} & \textbf{74.69} & \textbf{72.75} &  \textbf{70.64}  \\ \hline
% \end{tabular}
% }
% \label{tb:ablationTable}
% \end{table}

\begin{table}[]
\centering
\caption{Ablation study results in terms of PR-AUC.}
\vspace{-0.1in}
\resizebox{0.9\columnwidth}{!}{
\begin{tabular}{l|ccc|c}
\hline
\textbf{Dataset}  & \textbf{Kidney}   & \textbf{COPD}  & \textbf{Amnesia} & \textbf{MIMIC} \\ \hline
AS-1 & 76.91 & 70.19 & 71.90  & 67.98\\
AS-2 & 77.18 & 71.44 & 72.90 & 68.42\\
AS-3 & 76.82 & 71.00 &  69.17& 65.06\\ 
\ours & \textbf{77.88} & \textbf{72.03} & \textbf{74.69} &  \textbf{70.64}  \\ \hline
\end{tabular}
}
\vspace{-0.2in}
\label{tb:ablationTable}
\end{table}

% To examine the effectiveness of the key components of our model, we conduct the following ablation studies:
% \begin{itemize} 
% \item \textbf{Without using the hidden state $\mathbf{h}_{k-1}$ in the diffusion model} (\emph{AS-1}): When generating synthetic data $e_k^{\prime}$ in Section~\ref{sec:diffusion_model}, we do not consider the influence from the previous hidden state and remove $\mathbf{h}_{k-1}$ from Eq. (\ref{backward4}).
% \item \textbf{Without using the step-wise attention mechanism in Section~\ref{sec:step-wise-attention}} (\emph{AS-2}): We assign equal weights (0.5) to $\gamma_1$ and $\gamma_2$ in Eq. (\ref{fuse}).
% \item \textbf{Removing the regularization on the generated data} (\emph{AS-3}): We set $\lambda_S = 0$ by removing the regularization on the generated sequence in the loss function, i.e., Eq. (\ref{totalloss}).
% \end{itemize}

To examine the effectiveness of the key components of our model, we conduct the following ablation studies. \textbf{AS-1}: Without using the hidden state $\mathbf{h}_{k-1}$ in the diffusion model. When generating synthetic data $e_k^{\prime}$ in Section~\ref{sec:diffusion_model}, we do not consider the influence from the previous hidden state and remove $\mathbf{h}_{k-1}$ from Eq. (\ref{backward4}). \textbf{AS-2}: Without using the step-wise attention mechanism in Section~\ref{sec:step-wise-attention}. We assign equal weights (0.5) to $\gamma_1$ and $\gamma_2$ in Eq. (\ref{fuse}).
\textbf{AS-3}: Removing the regularization of the generated data. We set $\lambda_S = 0$ by removing the regularization on the generated sequence in the loss function, i.e., Eq. (\ref{totalloss}).

Table \ref{tb:ablationTable} presents the ablation study results. When components are removed from the model, there's a decline in PR-AUC scores across all datasets. In AS-1, removing the hidden state $\mathbf{h}_{k-1}$ in Eq. \eqref{fuse} that accounts for prior visits leads to performance drops, like a 2.6\% dip in the COPD dataset. Omitting past data, our model underperforms some baselines in Table \ref{resultTable1}, highlighting the importance of historical information in synthetic EHR data generation. In AS-2, by setting both attention weights $\gamma^e_n$ and $\gamma^h_n$ to 0.5, the model's attention to the previous visit reduces, causing a performance decrease. However, it still accesses information from $\mathbf{h}_{k-1}$. Lowering attention to $\mathbf{h}_{k-1}$ consistently diminishes performance, emphasizing the need for prior information in EHR data augmentation. In AS-3, removing the regularization term by setting $\lambda_S$ to 0 results in a performance drop due to unrestricted noise from synthetic data affecting predictions. Conclusively, the ablation study validates the necessity of specific data generation mechanisms for health risk prediction, and every module in our \ours is essential for optimal performance.

\begin{table}[]
\caption{Performance of GAN-based generators.}
\vspace{-0.1in}
\centering
\resizebox{0.9\columnwidth}{!}{
\begin{tabular}{c|c|c|c|c}
\hline
\textbf{Datasets}     & \textbf{Kidney} & \textbf{COPD} & \textbf{Amnesia} & \textbf{MIMIC} \\ \hline
% Metric       &    PR-AUC    &   PR-AUC      &        PR-AUC       &       PR-AUC    \\ \hline
LSTM         & 61.07 & 55.34 & 53.63  & 59.43\\\hline
ehrGAN       &    68.10 & 64.54 & 64.64 & 57.28\\
GcGAN        &    68.94 & 65.79 & 64.72 & 58.46\\
actGAN       &    69.81 & 64.39 & 65.11 & 61.88\\
medGAN       &    70.00 & 65.78 & 64.32 & 62.81\\\hline
ehrGAN+Att  &72.22&68.66&70.87&58.47\\
GcGAN+Att   &72.67&67.04&70.39&61.27\\
actGAN+Att  &72.51&68.17&69.72&62.74\\
medGAN+Att  &71.72&67.46&69.65&63.06\\
\ours & \textbf{77.88} & \textbf{72.03} & \textbf{74.69} & \textbf{70.64}\\ \hline
\end{tabular}
}
\vspace{-0.2in}
\label{exp3}
\end{table}

\subsection{Comparison Against GAN-based Generators}

In this experiment, we assess the end-to-end prediction and step-wise attention strategies on GAN-based EHR generation models. Current GAN-based EHR generation methods~\cite{ehrGAN,GcGAN,actGAN,medGAN} are not directly compatible with risk prediction due to their task-unrelated design. Thus, we first fit these methods with our LSTM hidden state learner and classifier in \ours. Each GAN generator utilizes visit embeddings $\mathbf{e}_k$ for synthetic visits and hidden state embeddings. In the first part of the experiment, the original hidden state $\mathbf{h}_k$ is not used, and we later introduce the Step-wise Attention Mechanism to all baselines, labeled as (+Att). Results in Table \ref{exp3} indicate that GAN-based methods see an improved performance against the LSTM baseline and even more with the attention mechanism. Moreover, \ours consistently surpasses all GAN-based methods, reinforcing our model's superiority in data augmentation and prediction. The step-wise attention method proves especially potent for data with significant time dependencies, underlining its role in maintaining data sequence integrity.

% For each baseline model, we use its visit embedding $\mathbf{e}_k$ and hidden state $\mathbf{h}_k$ obtained via RNN or Transformer as the input for the diffusion module and step-wise attention to produce synthetic data embedding $\mathbf{e}_k^\prime$ and synthetic hidden state $\mathbf{h}_k^\prime$. Then the synthetic hidden state $\mathbf{h}_k^\prime$ is forwarded to subsequent modules for prediction. According to Eq.\eqref{totalloss}, all baseline models with diffusion carry additional loss elements for the prediction of generated data and the diffusion module, and we also only use $\hat{y}$ from the original data sequence as the prediction result.

% We perform each experiment five times and reported the mean metric values. We illustrate our results in Figure \ref{exp2}, which displays paired bar plots of six baseline models across four datasets, with PR-AUC as the benchmark. The blue bars signify primary models with the diffusion module and step-wise attention, whereas the yellow bars denote models without these. It is evident that the proposed diffusion and step-wise attention methods considerably enhance performance across almost all datasets and baselines, thereby reinforcing their importance and usefulness.

\subsection{Synthetic EHR Data Analysis}
We follow ehrGAN~\cite{ehrGAN} to analyze the synthetic EHR data generated by the proposed {\ours}. In particular, we use the trained $\mathbf{W}_v$ in Eq.~\eqref{codeEmbedding} as the lookup dictionary, which is the learned ICD code embeddings. We feed each single ICD code $c_i$ into our model to produce a corresponding visit representation $\mathbf{e}^\prime_i$ using Eq.~\eqref{backwardgamma}, which can be treated as the generated ICD code embedding. We then calculate the cosine similarity between $\mathbf{e}^\prime_i$ and each code embedding in $\mathbf{W}_v$, and the ICD code with the highest cosine similarity can be considered as the mapping of the synthetic code. Finally, we count the number of mapped codes in the synthetic data and compare them with the frequency distribution of the original dataset.

% In order to produce human-readable ICD code distribution, we first utilize our trained model's embedding layers and combined test, train, and validation datasets  to generate code-embedding mapping as lookup dictionaries, per target disease. We then utilize standalone ICD codes from the test set to produce synthetic ICD embeddings, at a 1:1 ratio. Lastly, we calculate the cosine similarity between each generated embedding and all embeddings in the lookup table, and choose the ICD code with the top 1 highest similarity score as the code mapping result. We utilize this strategy in all the following code distribution experiments and illustrations.

In Table \ref{FreqTable:SynAmnesia}, we highlight the top 10 ICD codes most commonly found in the synthetic dataset of Amnesia ($\mathcal{D}_g$) and also display their ranks within the original dataset($\mathcal{D}_o$). Impressively, our model has unearthed lesser-known risk factors, elevating their prominence within the synthetic dataset.

The ICD code ``287.30'' corresponds to ``Primary thrombocytopenia, unspecified'', commonly known as immune thrombocytopenic purpura (ITP)~\cite{liu2023we}. This immune-mediated bleeding disorder results in autoantibodies destroying a patient's platelets. Concurrently, amnesia or memory loss is a frequent symptom of dementia. While on the surface, there may seem to be no overt connection between the two, the research~\cite{ahn2002vascular} has illustrated that memory loss or amnesia is a frequent initial complaint among ITP patients. In certain instances, this condition can swiftly evolve into dementia. This hidden risk factor, initially ranked 3,234 in the dataset, has been given significant importance by our model in the generated dataset, and its rank raised to the top spot. This analysis shows that our proposed model has successfully captured hidden risk factors of the target disease and gives them significant attention in the generated dataset, while it is possible to interpret generated data in a human-readable format. 
% In the COPD dataset, the ICD code ``368.40'' stands for Visual field defect, unspecified''. At first glance, there also might not be an apparent link between this and the primary ailment of COPD. Nonetheless, investigation~\cite{mikaeili2015correlation} has revealed that COPD, predominantly a respiratory disease, can precipitate hypoxemia - a condition characterized by reduced blood oxygen levels. Prolonged hypoxemia can result in peripheral nerve damage, including the optic nerve, subsequently manifesting as visual field defects. The rank of code ``368.40'' raises from 2092 in the original dataset to 3 in the synthetic dataset. 

\begin{table}[t]
\vspace{-0.1in}
\centering
\caption{Top 10 most frequent ICD codes in the synthetic Amnesia dataset.}
\label{FreqTable:SynAmnesia}
\resizebox{0.95\columnwidth}{!}{
\begin{tabular}{llll}
\hline
\textbf{R($\mathcal{D}_g$)} & \textbf{R($\mathcal{D}_o$)} & \textbf{ICD-9} & \textbf{Descriptions} \\ 
\hline
1&	3234&	287.30& \makecell[l]{Primary thrombocytopenia, unspecified}\\ \hline
2&	3480&	V71.5& \makecell[l]{Observation following alleged\\ rape or seduction}\\ \hline
3&	3130&	493.11& \makecell[l]{Intrinsic asthma with status  asthmaticus}\\ \hline
4&	4847&	622.4& \makecell[l]{Stricture and stenosis of cervix}\\ \hline
5&	2567&	732.4& \makecell[l]{Obstetrical blood-clot embolism,\\ postpartum condition or complication}\\ \hline
6&	3607&	714.4& \makecell[l]{Chronic postrheumatic arthropathy}\\ \hline
7&	2882&	718.97& \makecell[l]{Unspecified derangement of joint,\\ ankle and foot}\\ \hline
8&	31&	309.81& Posttraumatic stress disorder\\ \hline
9&	1988&	V82.89& \makecell[l]{Special screening for other specified \\conditions}\\ \hline
10&	1941&	719.00& Effusion of joint, site unspecified\\ \hline

\end{tabular}
}
\vspace{-0.1in}
\end{table}
\section{Conclusion}
We present \ours, a novel diffusion-based health risk prediction model with data augmentation. It captures temporal relationships by visit-level time embedding and hidden states, while the diffusion module creates synthetic data based on current and past information with an attention mechanism. Tested on four real-world datasets, it outperforms existing models. Further experiments reinforce its validity, while synthetic data analysis reflects the model's interpretability. Future endeavors will adapt it for multi-modal data in a broader predictive framework.
{\small
\bibliographystyle{plain}
\bibliography{bibfile}
}

\newpage

\section{Appendix}

\subsection{Training Algorithm}
Algorithm \ref{alg:cap} illustrates the training algorithm of \ours with source code.

\begin{algorithm}[]
\caption{Training Procedure (without data batches)}\label{alg:cap}
\begin{algorithmic}[1]\small
\REQUIRE Training data $\mathrm{D}_{tra} = \{V_1, \cdots, V_J\}$, validation set $\mathrm{D}_{val}$, hyperparameters, and learning rate
\ENSURE Trained model parameter $\theta_{best}$
\STATE Initialize \textit{ours} with random parameter $\theta$;
\FOR {epoch in EPOCH}
\FOR {patient $j =1$ to $J$}  
    \STATE \textcolor{blue}{//Original EHR data}
    \FOR{visit $k = 1$ to $K$}
        \STATE Obtain visit embedding $\mathbf{e}_k$ according to Eq.~\eqref{eq:visit_embedding};
        \STATE Generate hidden state $\mathbf{h}_k$ according to Eq.~\eqref{eq:hidden_states};
    \ENDFOR
    \STATE \textcolor{blue}{//Diffusion-based data augmentation}
    \FOR{visit $k = 1$ to $K$}
        \FOR{step $n=0$ to $N$}
        \STATE Calculate $\gamma_e^n$ and $\gamma_h^n$ according to Eq.~\eqref{fuse};
        \STATE Generate data sample $\mathbf{e}_{k,n}^\prime$ according to \eqref{backwardgamma};
        \ENDFOR
        \STATE Generate synthetic hidden state $\mathbf{h}_k^\prime$ according to \eqref{eq:hidden_states};
    \ENDFOR
    \STATE \textcolor{blue}{//Risk prediction}
    \STATE Calculate prediction results $\hat{\mathbf{y}}_j$ and $\hat{\mathbf{y}}^\prime_j$ according to Eq.~\eqref{eq:prediction} and Eq.~\eqref{eq:prediction_fake};
    \STATE Update the loss value;
\ENDFOR
\STATE \textcolor{blue}{//Loss optimization}
\STATE Update parameters $\theta$ by optimizing the loss $\mathcal{L}$ according to Eq.~\eqref{totalloss};
\STATE \textcolor{blue}{//Model selection}
\STATE Calculate $F1$ Score on validation set $\mathrm{D}_{val}$;
\IF{${\text{F1}}_v \leq {\text{F1}}_v^{\text{min}}$}
    \STATE $\theta_{best}$ = $\theta$;
\ENDIF
\ENDFOR
\end{algorithmic}
\end{algorithm}

\begin{figure}[t]
\centering
    \includegraphics[width=0.9\columnwidth]{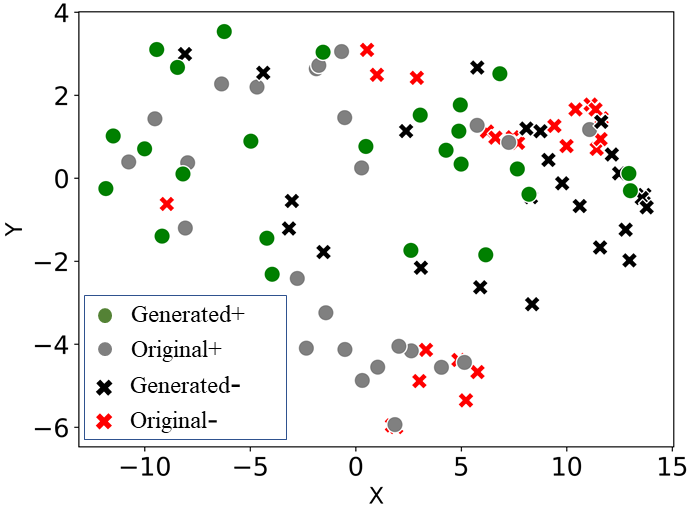}
    \caption{Visualization of synthetic and original embeddings of two categories with tSNE on the kidney dataset.}
    \label{embeddingPlot2}
    \vspace{-0.1in}
\end{figure}

\begin{figure*}[t]
    \centering
    \includegraphics[width=0.9\textwidth]{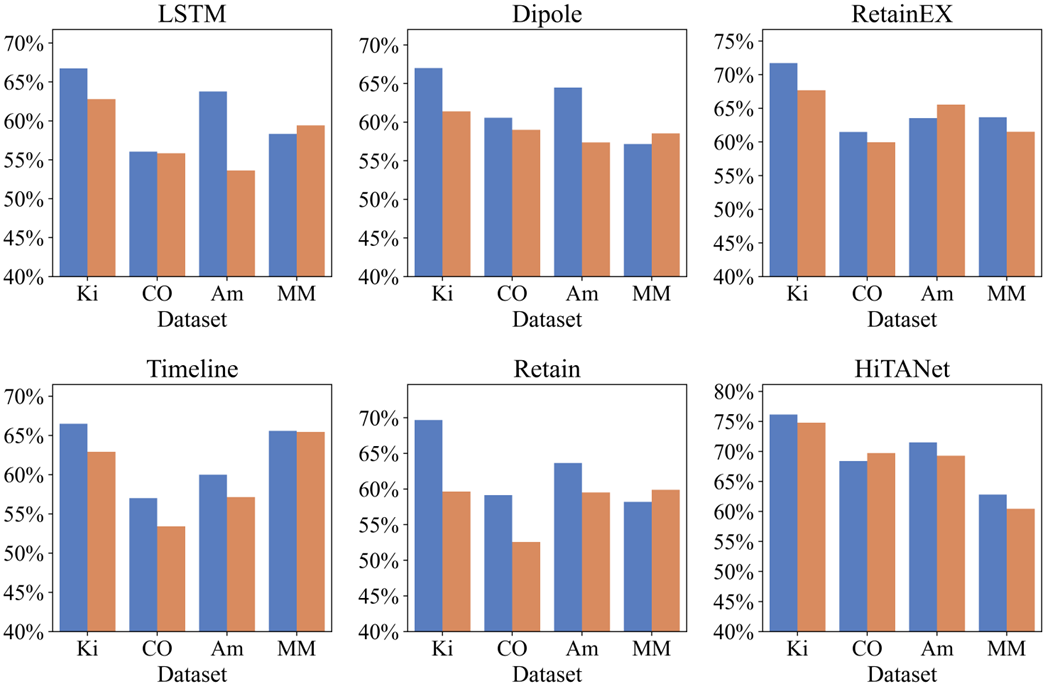}
    \caption{Performance validation for baselines with (blue) or without (yellow) adding diffusion data augmentation on all datasets, where ``Ki'' for Kidney, ``CO'' for COPD, ``Am'' for Amnesia, ``HF'' for Heart Failure, and ``MM'' for MIMIC on $x$-axis.}
    \label{exp2}
\end{figure*}

\subsection{Baseline Explanation}
In this work, we utilize the following health risk prediction models as baselines. LSTM~\cite{lstm} is the basic model, and Dipole~\cite{dipole} and Retain~\cite{retain} are relatively early methods of RNN and use attention mechanisms. Adacare~\cite{adacare} utilizes multiple-scale dilated convolutions to extract temporal relationships on different scales. Integrating time information, Timeline~\cite{timeline}, RetainEx~\cite{retainex}, and T-LSTM~\cite{tlstm} either model the time decay or separate patients into similar characteristics by time information. MedSkim~\cite{medskim} is equipped with a code and visit selection method to filter out unrelated information. SAnD~\cite{sand}, LSAN~\cite{lsan}, and HiTANet~\cite{hitanet} are Transformer-based methods that generate visit and code attention on different scales. 
We also include three \textbf{EHR augmentation} models as baselines.
MaskEHR~\cite{ma2020rare} focuses on predicting risks for rare diseases, which only generates synthetic data for positive cases using generative adversarial networks and reinforcement learning techniques. 
ehrGAN~\cite{ehrGAN} first utilizes an encoder-decoder CNN structured GAN to learn and generate EHR data and then trains a CNN predictor separately. 
TabDDPM~\cite{kotelnikov2023tabddpm} is a diffusion-based model, which utilizes MLP blocks structured diffusion model to generate both numerical and categorical EHR data and train a separate LSTM classifier as the risk prediction module.

\subsection{Original vs Synthetic: Hidden States Analysis}

To assess the efficacy of our model on individual samples' hidden state, we take a random selection of 50 data points from the Kidney dataset. From these samples, we record the hidden states, represented as $\mathbf{h}_k$ from the original data and $\mathbf{h}_k'$ from the synthetic data, corresponding to the last $K$-th visit $\mathbf{v}_K$.

To better visualize these high-dimensional hidden states, we utilize t-distributed stochastic neighbor embedding (tSNE). This technique enables us to project the data into a two-dimensional space, as shown in Figure \ref{embeddingPlot2}. In this representation, the X-axis and Y-axis signify the mapped values of each dimension, respectively. In terms of notation, we label $\mathbf{h}_k'$ derived from a positive sample as ``Generated+'', and when derived from a negative sample as ``Generated-''. Conversely, $\mathbf{h}_k$ from a positive sample is marked as ``Original+'' and from a negative sample as ``Original-''. For further clarity in visual representation, we use circles to signify positive samples and crosses for negative samples.

Upon reviewing Figure \ref{embeddingPlot2}, a notable observation is that the synthetic or generated data clusters closely around the original, authentic Electronic Health Record (EHR) data. This suggests a promising degree of similarity between them. Additionally, the subtle shift in class boundaries when increasing the data count within each category, using our diffusion model, indicates the model's capability to augment data. This reinforces the notion that our model, \ours, has an efficient data augmentation capability.

% \subsection{Step-wise Attention Mechanism Visualization}

% In Figure \ref{gamma_continued}, we show the value distribution of $\gamma_K^e$ from the remaining Kidney, Amnesia, and COPD datasets. For these datasets, the model tends to mainly utilize previous visit information by producing a lower value of $\gamma_K^e$. On the contrary, the model tends to produce high $\gamma_K^e$ on the MIMIC dataset and gives heavy emphasis on the current visit's information. This could be due to the fact that the MIMIC dataset has 2.61 average visits per patient, which is significantly less than that of other datasets, and it shows that our proposed attention mechanism can adapt to different characteristics of the input datasets.

% \begin{figure*}[h]
% \centering

% \begin{subfigure}{0.9\columnwidth}
% \centering
% \includegraphics[width=1\linewidth]{figures/Amnesia.png}

% \end{subfigure}
% \hfill
% \begin{subfigure}{0.9\columnwidth}
% \centering
% \includegraphics[width=1\linewidth]{figures/COPD.png}

% \end{subfigure}

% \vspace{1em}

% \begin{subfigure}{0.9\columnwidth}
% \centering
% \includegraphics[width=1\linewidth]{figures/Kidney.png}

% \end{subfigure}
% \hfill
% \begin{subfigure}{0.9\columnwidth}
% \centering
% \includegraphics[width=1\linewidth]{figures/MIMIC.png}

% \end{subfigure}
% \vspace{-0.1in}
% \caption{Value distribution of $\gamma_K^e$ from selected epochs}
% \label{gamma_continued}
% \end{figure*}

\subsection{Equipping Baselines with the Proposed Diffusion Generation Module}

The proposed \ours is able to be integrated with various baseline models to notably elevate their predictive performance. To validate this claim, we conduct the following experiments.

We use visit embeddings $\mathbf{e}_k$ and hidden states $\mathbf{h}_k$ from baseline models as inputs for diffusion and step-wise attention to create synthetic data. These synthetic states $\mathbf{h}_k^\prime$ are then used for prediction, incorporating additional loss elements like in Eq.\eqref{totalloss}. Only predictions from original data sequences are reported. We perform each experiment five times and averaged the metrics, displayed in Figure \ref{exp2}. The figure shows that models with the diffusion module and step-wise attention (blue bars) consistently outperform those without (yellow bars) across multiple datasets, highlighting their effectiveness.

% \begin{figure}[h]
% \centering
% \includegraphics[width=0.9\columnwidth]{figures/Amnesia.png}
% \includegraphics[width=0.9\columnwidth]{figures/COPD.png}
% \includegraphics[width=0.9\columnwidth]{figures/Kidney.png}
% \includegraphics[width=0.9\columnwidth]{figures/MIMIC-III.png}
% \caption{Value distribution of $\gamma_K^e$ from selected epochs}
% \label{gamma_continued}
% \end{figure}

\begin{figure*}[p]
    \centering
    \includegraphics[width=0.9\textwidth]{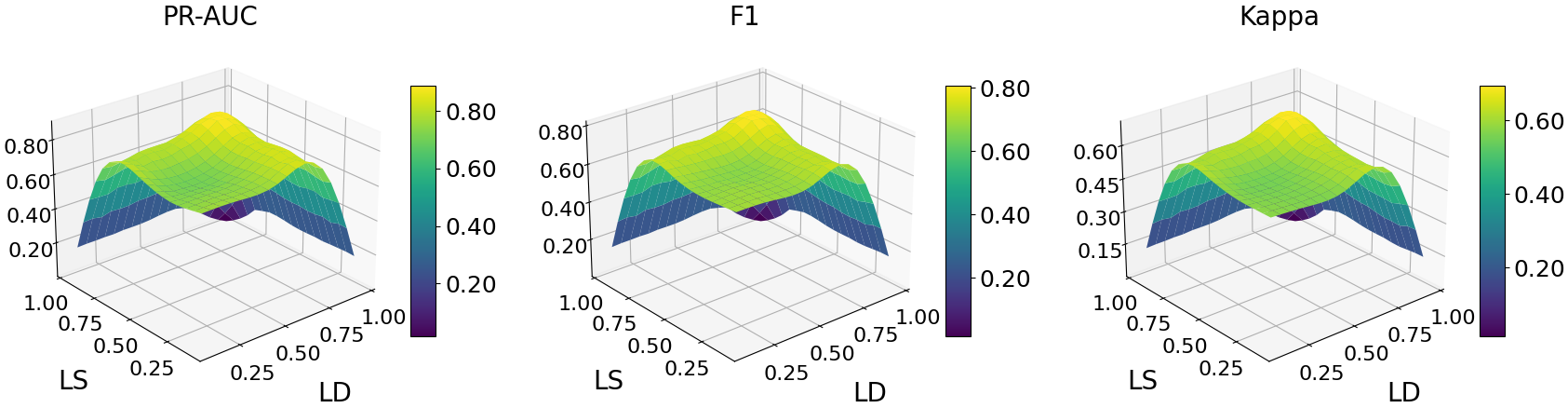}
    \caption{Hyper-parameter study visualization on the Kidney dataset}
    \label{fig:Kidney_hyper}
        \centering
    \includegraphics[width=0.9\textwidth]{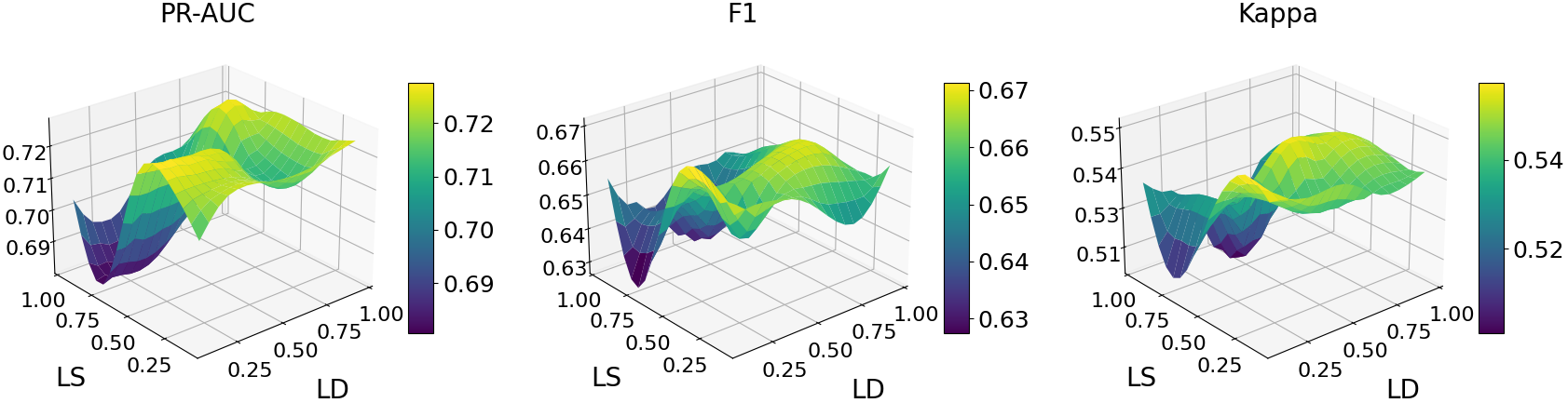}
    \caption{Hyper-parameter study visualization on the COPD dataset}
    \label{fig:COPD_hyper}
        \centering
    \includegraphics[width=0.9\textwidth]{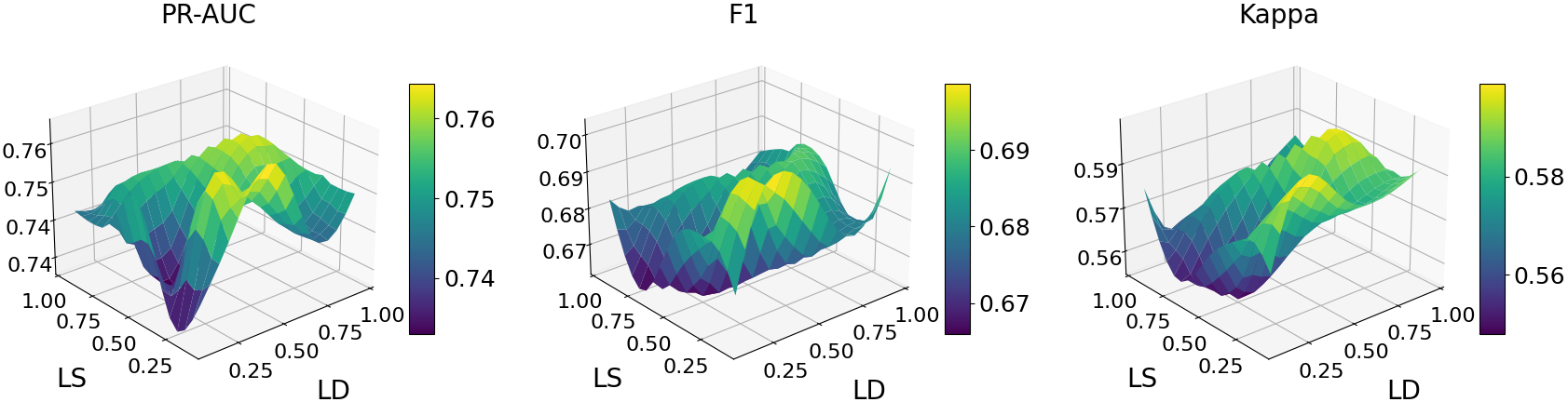}
    \caption{Hyper-parameter study visualization on the Amnesia dataset}
    \label{fig:Amnesia_hyper}
        \centering
    % \includegraphics[width=0.9\textwidth]{figures/Heart_failure_hyper.png}
    % \caption{Hyper-parameter study visualization on the Heart Failure dataset}
    % \label{fig:Heart_failure_hyper}
    %     \centering
    \includegraphics[width=0.9\textwidth]{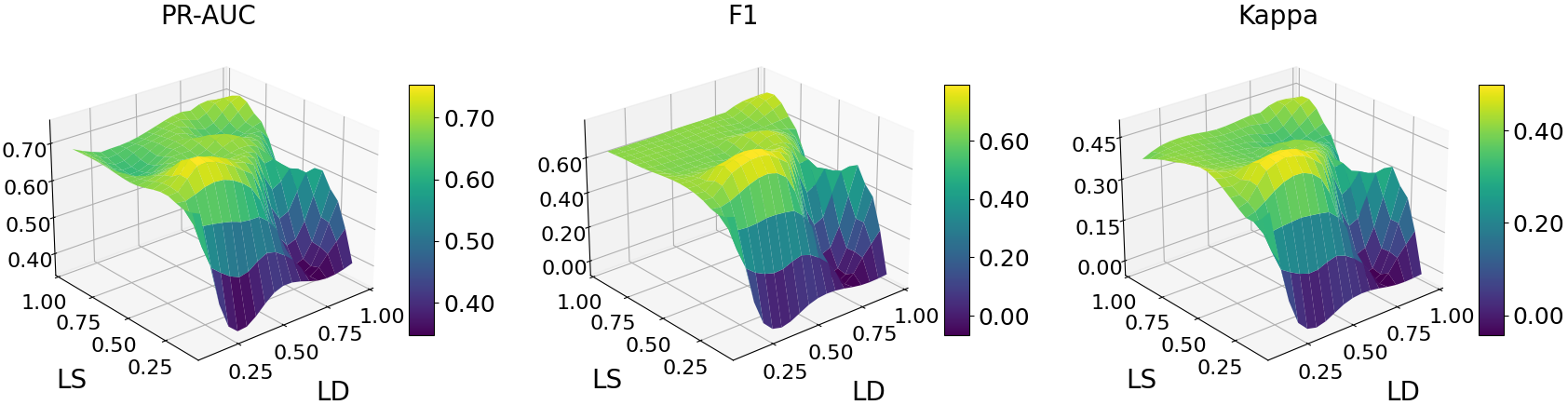}
    \caption{Hyper-parameter study visualization on the MIMIC dataset}
    \label{fig:mimic_hyper}
\end{figure*}

\subsection{Hyperparameter Study}
In this section, we explore the effect of hyperparameters on the overall performance of the proposed \ours. $\lambda_D$ controls the magnitude of the diffusion loss, giving the model a restriction on how much deviation is kept in the diffusion; $\lambda_S$ controls the relative importance of the generated sequence and punishes the model if the generated sequence does not produce the same prediction as the original prediction. We set up a grid of possible values of $\lambda_D=[0.1,0.25,0.5,0.75, 1.0]$ and $\lambda_S=[0.1,0.25,0.5,0.75, 1.0]$, and for each pair, we train our model on the all datasets and record performance in terms of PR-AUC, F1, and Kappa. We visualize the results in the landscape format, as shown in Figure \ref{fig:Kidney_hyper}, \ref{fig:COPD_hyper}, \ref{fig:Amnesia_hyper}, and \ref{fig:mimic_hyper} accordingly.

\end{document}